\documentclass[conference,a4paper]{IEEEtran}
\IEEEoverridecommandlockouts

\usepackage[hidelinks]{hyperref}
\usepackage[cmex10]{amsmath}
\usepackage{amssymb,amsfonts}
\usepackage{dblfloatfix}

\usepackage[ruled,vlined]{algorithm2e}
\usepackage{graphicx}
\graphicspath{{Figures/PDF/}{Figures/PNG/}}

\usepackage{booktabs}
\usepackage{siunitx}
\usepackage[numbers,compress]{natbib}
\usepackage{texnames}
\usepackage{bm,bbm}
\usepackage{orcidlink}
\usepackage{subcaption}
\usepackage{multirow}

\begin{document}
\title{SEMANTIC-AWARE SHIP DETECTION WITH VISION-LANGUAGE INTEGRATION}

\author{\IEEEauthorblockN{Jiahao Li}
	\IEEEauthorblockA{
		\textit{Department of } \\
		\textit{Earth System Science,} \\
		\textit{Tsinghua University}\\
		100084 Beijing, China\\
		lijiahao23@mails.tsinghua.edu.cn}
	\and
	\IEEEauthorblockN{Jiancheng Pan}
	\IEEEauthorblockA{
		\textit{Department of } \\
		\textit{Earth System Science,} \\
		\textit{Tsinghua University}\\
		100084 Beijing, China\\
		jiancheng.pan.plus@gmail.com}
	\and
	\IEEEauthorblockN{Yuze Sun}
	\IEEEauthorblockA{
		\textit{Department of } \\
		\textit{Earth System Science,} \\
		\textit{Tsinghua University}\\
		100084 Beijing, China\\
		syz23@mails.tsinghua.edu.cn}
	\and
	\IEEEauthorblockN{Xiaomeng Huang*}
	\IEEEauthorblockA{
		\textit{Department of } \\
		\textit{Earth System Science,} \\
		\textit{Tsinghua University}\\
		100084 Beijing, China\\
		hxm@tsinghua.edu.cn }
}
\vspace{-15mm}

\maketitle
\begin{abstract}
Ship detection in remote sensing imagery is a critical task with wide-ranging applications, such as maritime activity monitoring, shipping logistics, and environmental studies. However, existing methods often struggle to capture fine-grained semantic information, limiting their effectiveness in complex scenarios. To address these challenges, we propose a novel detection framework that combines Vision-Language Models (VLMs) with a multi-scale adaptive sliding window strategy. To facilitate Semantic-Aware Ship Detection (SASD), we introduce ShipSem-VL, a specialized Vision-Language dataset designed to capture fine-grained ship attributes. We evaluate our framework through three well-defined tasks, providing a comprehensive analysis of its performance and demonstrating its effectiveness in advancing SASD from multiple perspectives.
\end{abstract}


\section{Introduction}
\label{sec:intro}

Ship detection in remote sensing imagery is a critical task in the field of Earth sciences, playing an essential role in monitoring maritime activities, analyzing carbon emissions from sea traffic, managing shipping logistics, and supporting meteorological studies~\cite{paolo2024satellite}. Currently, mainstream approaches for ship detection typically rely on object detection frameworks, such as Faster R-CNN, YOLO, and U-Net, which are trained on annotated datasets~\cite{ren2016faster,du2018understanding}. While existing datasets account for different ship types, such as fishing vessels and cargo ships, they often lack sufficient focus on the finer details of ships~\cite{paolo2024satellite}. In many practical scenarios, the need to track specific ships—such as identifying vessels based on their hull numbers, colors, or unique markers—requires a high level of detail and precision. This has led to the formulation of a new task, Semantic-Aware Ship Detection (SASD), which focuses on leveraging semantic descriptions to characterize specific ship attributes and enable precise detection.

\begin{figure}[bt]
	\centering
	\includegraphics[width=\linewidth]{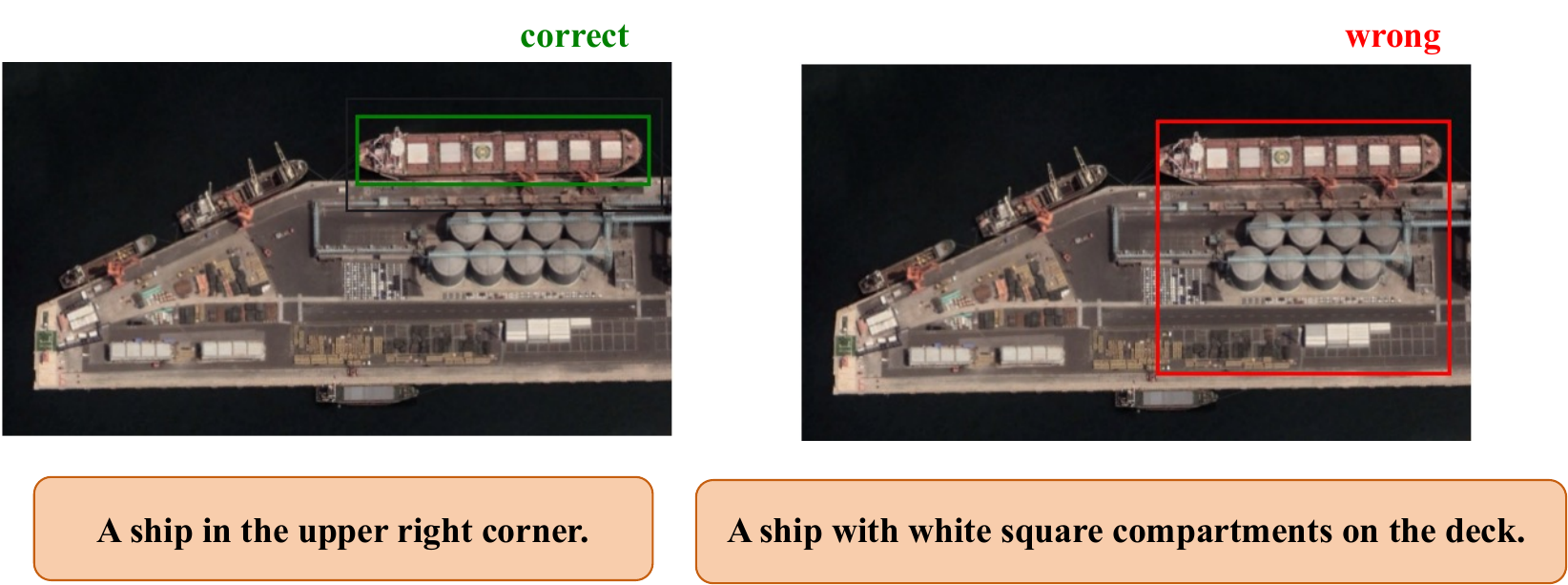}
	\caption{Grounding models in remote sensing still have gaps in understanding the details of ships.}\label{fig:fig1}
\end{figure}

The development of Vision-Language models offers new approaches for SASD task. By leveraging detailed semantic descriptions, it is possible to perform specific ship detection and tracking, elevating the task of ship detection to a more advanced level.

We propose an advanced ship detection method that leverages VLMs in combination with an adaptive sliding window strategy, supported by a dedicated Vision-Language dataset, ShipSem-VL. To comprehensively evaluate the accuracy and robustness of our method, we further subdivided the SASD task into three detailed perspectives, designing 45 specific tasks that encompass diverse ship types and varying scene complexities, ensuring a thorough assessment across multiple dimensions.

In conclusion, the contributions of this paper are summarized as follows.
\begin{enumerate}
	\item We constructed a Vision-Language dataset specifically designed for SASD, ShipSem-VL. ShipSem-VL encompasses a wide range of ship types across the globe and includes detailed semantic annotations;
	\item We propose an innovative semantic-aware ship detection framework that integrates an adaptive sliding window mechanism with text-image contrastive learning. Trained on the ShipSem-VL dataset, this approach effectively addresses the SASD task by enabling precise ship detection through semantic representation;
	\item We further divided the SASD task into three detailed perspectives to comprehensively evaluate the effectiveness of the dataset and method, providing a deeper understanding of their performance and potential areas for improvement in Vision-Language remote sensing detection.
\end{enumerate}

\begin{figure}[bt]
	\centering
	\includegraphics[width=\linewidth]{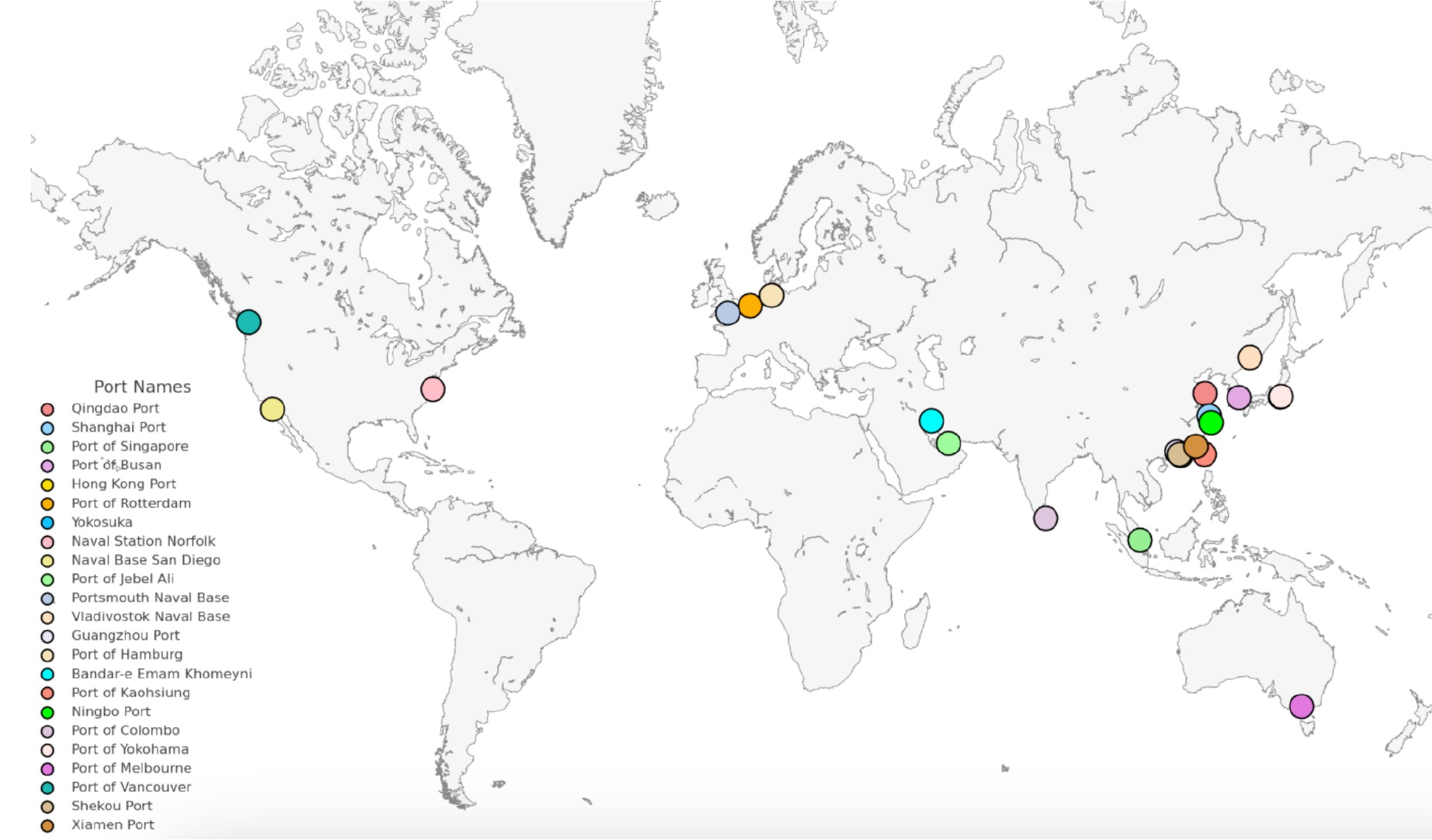}
	\caption{Sample distribution of the ShipSem-VL dataset.}\label{fig:fig2}
\end{figure}

\section{Related Works}
\label{sec:related}
CLIP, through contrastive learning on large-scale image-text pairs, has demonstrated exceptional generalization and transfer capabilities in open-domain scenarios~\cite{radford2021learning,shen2021much}. The core idea of CLIP is to learn a joint representation of images and text, enabling the model to identify relevant objects in unseen images based on semantic descriptions. Building on the foundational concept and architecture of CLIP, the remote sensing field has developed a series of models such as RemoteCLIP~\cite{liu2024remoteclip}, RS-CLIP~\cite{li2023rs}, PIR-CLIP~\cite{pan2023prior,pan2024pir}, and GeoRSCLIP~\cite{zhang2024rs5m}. These models, trained on multimodal remote sensing datasets, have shown promising results. However, their primary focus lies in the semantic understanding~\cite{pan2023reducing} of the overall scene, with limited attention to fine-grained detection of specific categories, such as ships.

Grounding tasks in the remote sensing domain share some similarities with our objectives, as they aim to locate regions in an entire image corresponding to a given semantic description. Methods like GLIP~\cite{li2022grounded}, RegionCLIP~\cite{zhong2022regionclip}, and RSVG~\cite{zhan2023rsvg} provide valuable insights for our research. Moreover, large language models (LLMs) fine-tuned on remote sensing datasets, such as RSGPT~\cite{hu2023rsgpt} and H2RSVLM~\cite{pang2024h2rsvlm}, also exhibit the ability to interpret remote sensing imagery. However, these models predominantly focus on holistic scene understanding and lack the capability to detect and localize fine-grained targets like ships.

We conducted experiments using the most advanced remote sensing grounding models available. As shown in Fig.\ref{fig:fig1}, these models, trained on datasets for overall remote sensing imagery, often prioritize the spatial relationships between ships and their surrounding environment. However, an overemphasis on ship-related semantic descriptions frequently leads to confusion within the model. Therefore, grounding tasks in the remote sensing domain cannot be directly equated with object detection tasks. Object detection focuses on the intrinsic characteristics of the target itself, while grounding emphasizes the spatial relationships within the entire remote sensing scene. This distinction highlights the importance of constructing a specialized Vision-Language dataset for ship detection as a crucial step toward advancing fine-grained detection research.

While advancements in grounding and related fields have provided valuable ideas for ship detection, the core focus of object detection differs significantly from grounding~\cite{sun2022visual,li2024vision}. Unlike grounding tasks, detecting specific targets requires a detailed understanding of the target’s fine-grained features and the ability to locate the target accurately in complex remote sensing imagery~\cite{ma2024direction}. Existing remote sensing grounding models~\cite{liu2024grounding,pan2025locate,pan2025enhance} are primarily trained on datasets representing the overall scene and lack fine-grained data specific to certain categories. As a result, the grounding process tends to emphasize spatial relationships between objects~\cite{liu2019improving}, often overlooking the detailed characteristics of fine-grained targets~\cite{li2024injecting}. In SASD task, current Vision-Language grounding methods are more inclined to focus on the spatial distribution of targets, while failing to adequately capture essential features such as color, shape, or distinctive markings.

\section{Methodology}
\label{sec:metho}

\subsection{ShipSem-VL Dateset}
Currently, Vision-Language datasets in the remote sensing field primarily focus on the semantic description of overall scenes. Such datasets enhance models' ability to understand the general context of scenes and facilitate the interpretation of spatial relationships within remote sensing imagery. However, these datasets are relatively limited when it comes to capturing fine-grained details of specific categories. In SASD task, more attention is required on the detailed characteristics of ship targets. To address this issue, we constructed a specialized semantic-image remote sensing dataset, ShipSem-VL, which provides detailed semantic descriptions of ships and their structural components across various scales.

\begin{figure}[bt]
	\centering
	\includegraphics[width=\linewidth]{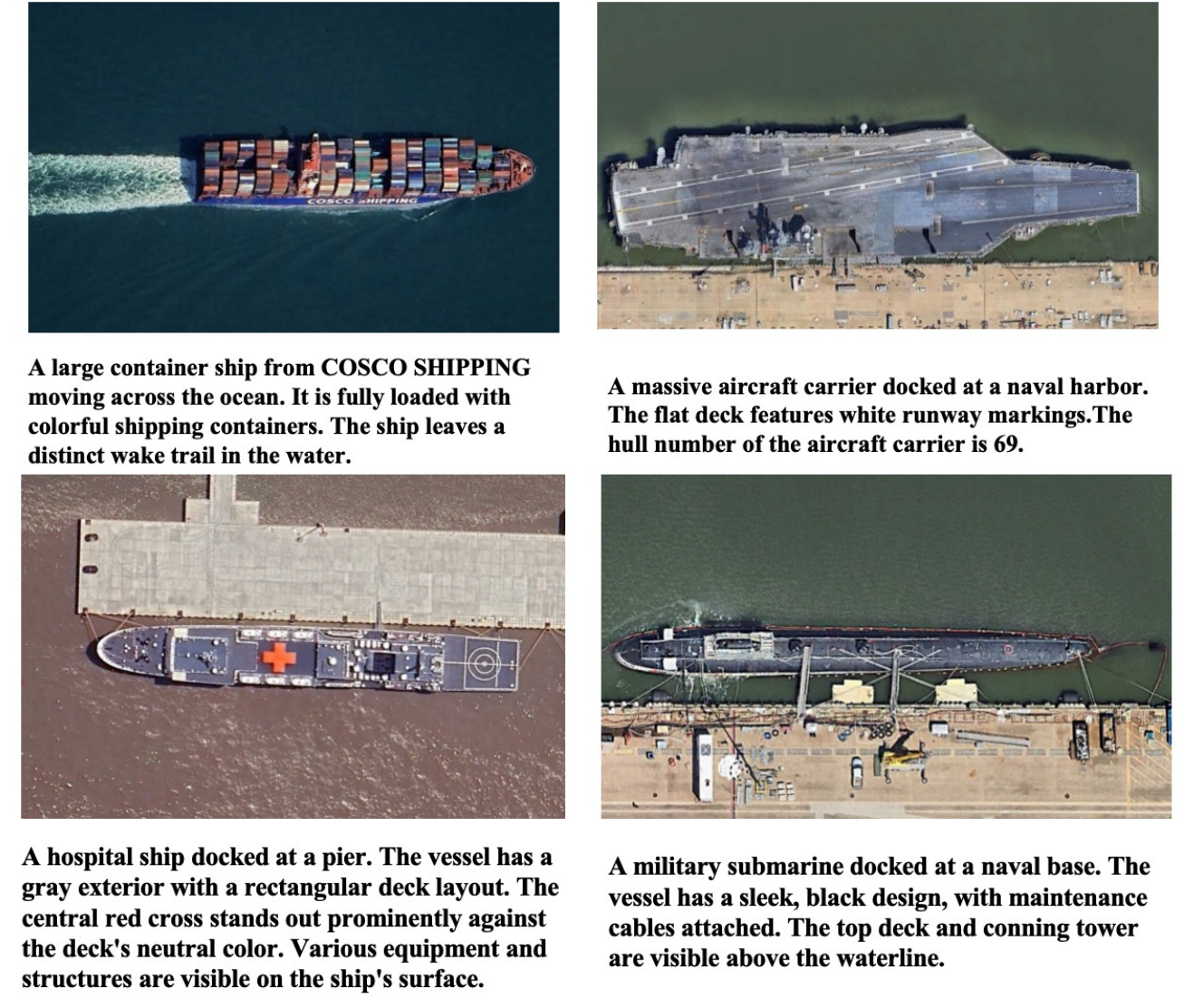}
	\caption{Examples of the ShipSem-VL dataset.}\label{fig:fig3}
\end{figure}

\begin{figure*}[t]
	\centering
	\includegraphics[width=0.75\linewidth]{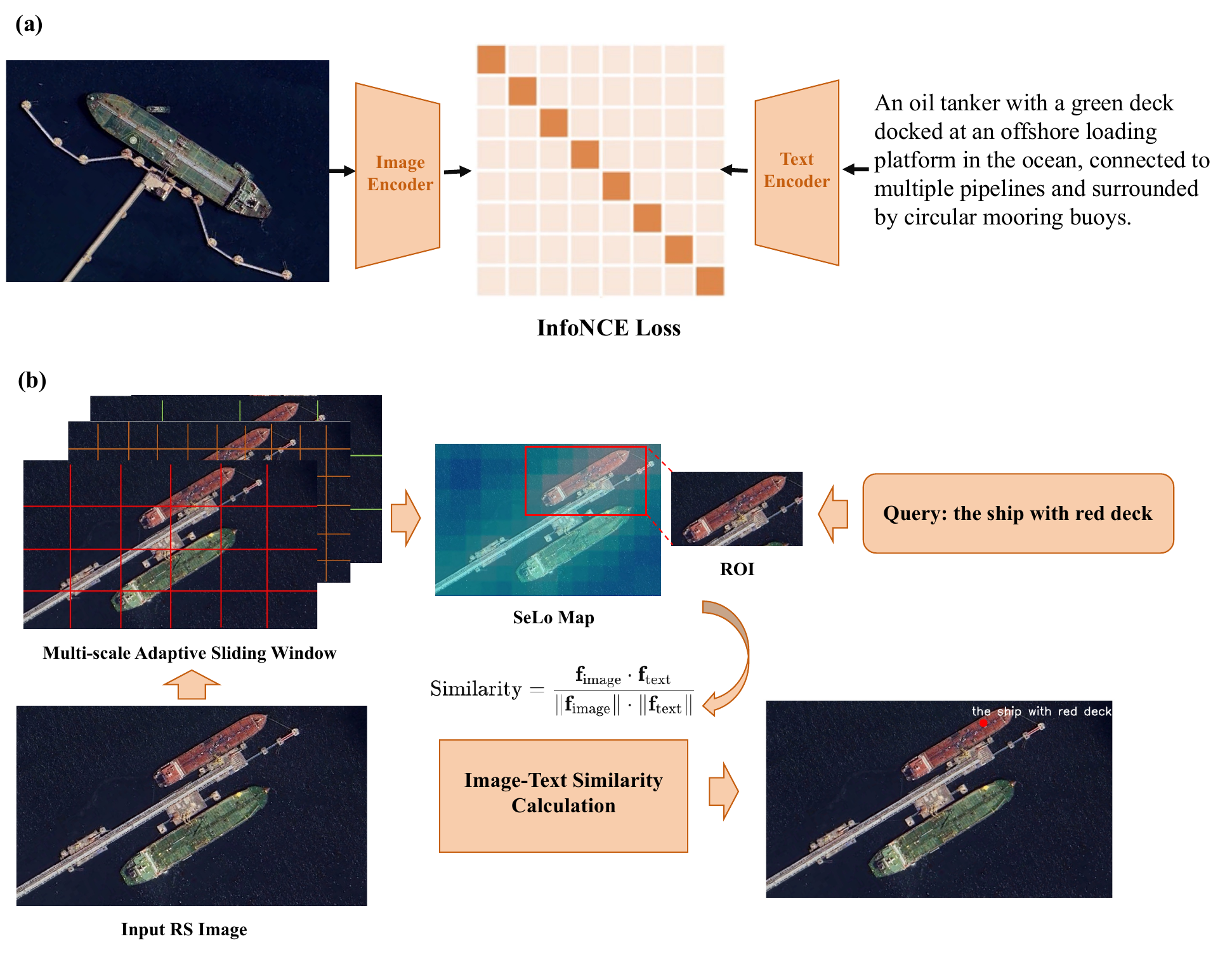}
	\caption{Framework for Semantic-Aware Ship Detection.}\label{fig:fig4}
\end{figure*}
The dataset was developed using imagery collected via Google Earth from 23 different types of ports and their surrounding sea areas worldwide (Fig.\ref{fig:fig2}), increasing the diversity of ship backgrounds while covering a wide range of ship types. All collected images underwent manual visual interpretation, followed by the addition of detailed semantic annotations, as shown in Fig.\ref{fig:fig3}. Unlike conventional datasets with precise location annotations, we aim to develop a method that relies solely on vision-language training, thereby reducing the workload associated with annotation. Our dataset does not include manually labeled bounding boxes for ship locations but instead consists of detailed textual descriptions of the ships. The final ShipSem-VL dataset contains 2,611 samples, offering a valuable resource for fine-grained ship detection tasks in remote sensing.

\subsection{Advanced Framework for Semantic-Aware Ship Detection}
\label{sec:metho.2}
Based on the ShipSem-VL dataset, we propose an innovative method for ship detection, as shown in Fig.\ref{fig:fig4}. Due to the need for higher descriptive detail in advanced ship detection, our approach differs from traditional object detection methods that rely on training with annotated bounding boxes. Specifically, we introduce a Vision-Language-based approach to achieve a deeper understanding of ship details. Our proposed method leverages a multi-scale sliding window strategy combined with a Vision-Language model, enabling fine-grained and precise detection of specific ship targets in remote sensing images.

Remote sensing images often exhibit targets with uneven resolution distribution, where the size and shape of ships can vary significantly due to imaging conditions. To address this challenge, we designed a multi-scale sliding window strategy. Initially, a Multi-scale Adaptive Sliding Window scan is applied to the entire image to identify high-confidence regions with high semantic similarity to the target description. These Regions of Interest (ROIs) are then further analyzed to refine detection accuracy.

For the underlying framework, we leverage the CLIP model as the foundation for remote sensing image-text comparison. We use the ViT-B-32 model as the initial pre-trained weights for fine-tuning on the ShipSem-VL dataset. The trained model is then utilized as the Image-Text Similarity Calculation module within our overall framework, significantly enhancing the performance of SASD task. The training process were conducted on 4 Sugon DCU (Deep Computing Unit), each featuring 16GB of memory. The batch size was set to 128 for each process. The training employed an initial learning rate of 1×$10^{-4}$, with a weight decay of 0.1 to regularize the model. A warmup phase consisting of 10,000 steps was applied to gradually increase the learning rate, stabilizing the optimization process in its early stages. These settings ensured efficient utilization of computational resources and facilitated stable convergence of the model during training.

\section{Results}
\label{sec:res}
To comprehensively evaluate the performance of our dataset and proposed method on the SASD task, we further divided the SASD task into three distinct perspectives and designed open experiments for each perspective. Following the experimental practices commonly adopted in VLMs research, we established an open ship semantic detection task and conducted manual evaluations of the experimental results.

\begin{figure}[bt]
	\centering
	\includegraphics[width=\linewidth]{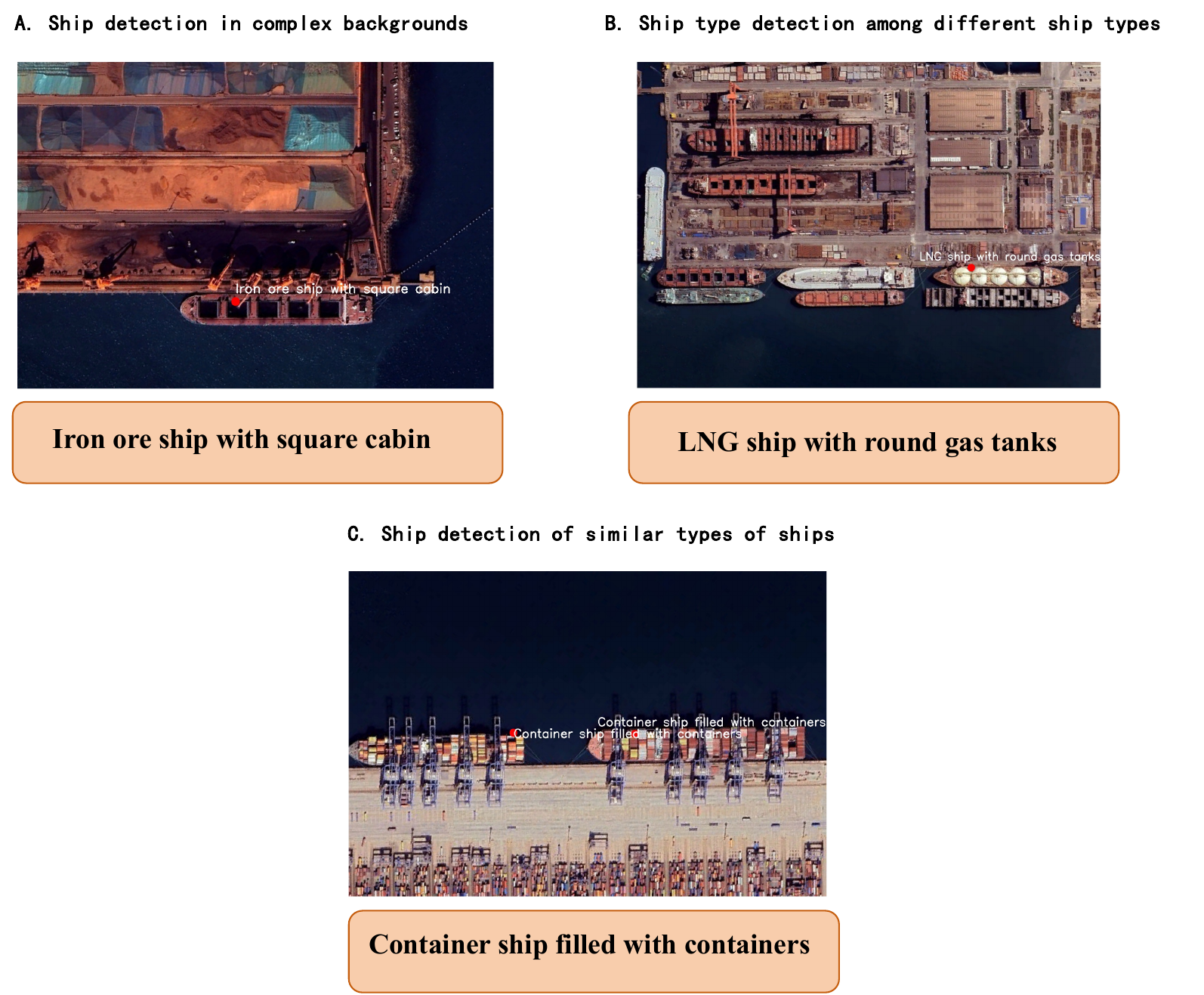}
	\caption{Specific examples and experimental results of task A.B.C.}\label{fig:fig5}
\end{figure}

We designed 45 specialized tasks, categorized into three types. Examples of the three task types are illustrated in the accompanying Fig.\ref{fig:fig5}.

\begin{enumerate}
	\item Task A: Detection of ships in complex backgrounds, designed to evaluate the model's ability to identify ships from remote sensing images.
	\item Task B: Detection of specific ship types based on complex semantic descriptions, designed to assess the model's understanding of ship-related semantic descriptions.
	\item Task C: Detection of visually similar ship types, where a single remote sensing image contains multiple targets that closely match the semantic description.
\end{enumerate}

The recall and precision results for the three task were shown in Table \ref{tab:performance-comparison}. The results show that ViT-ShipSemVL maintains a more stable balance between precision and recall across varying thresholds, indicating its robustness and adaptability in complex scenarios. The model achieves the most significant improvements in precision, particularly for Task B, where precision gains are essential due to the greater complexity of ship detection in challenging conditions. Interestingly, in Task C, a notable phenomenon was observed: as the threshold decreased, the recall rate also declined. Through a detailed breakdown and analysis of each step in our method, we discovered that this decline in recall was caused by misjudgments in regions adjacent to ships. Lowering the threshold led to the merging of neighboring ships with similar semantics into a single ROI during the sliding window process, ultimately reducing the total number of detected ship targets. While adopting a two-stage approach, such as using object detection algorithms like YOLO to extract ships first and then applying a vision-language model for semantic judgment, could potentially address this issue. However, a two-stage detection approach could introduce omissions due to missed detections in the first stage. Additionally, our primary objective is to develop a method that does not rely on annotated bounding boxes for training, thereby reducing labeling costs. Therefore, our proposed approach was carefully designed as a comprehensive solution to balance these considerations. The results of Task C suggest that careful selection of the threshold is crucial when applying our method. 
\begin{table}[ht]
\centering
\renewcommand{\arraystretch}{1.5} 
\caption{Recall and Precision results for three tasks.}
\label{tab:performance-comparison}
\large 
\resizebox{\linewidth}{!}{%
\begin{tabular}{llllllll}
\hline
\textbf{}    & \multicolumn{2}{c}{\textbf{Task A}} & \multicolumn{2}{c}{\textbf{Task B}} & \multicolumn{2}{c}{\textbf{Task C}} \\ \hline
\textbf{}    & \textbf{Recall} & \textbf{Precision} & \textbf{Recall} & \textbf{Precision} & \textbf{Recall} & \textbf{Precision} \\ \hline
\textbf{Threshold 0.8}  \\
ViT-B-32      & 76.62 & 80.00 & 80.00 & 63.16 & 87.09 & 77.14 \\
ViT-ShipSemVL & 84.61 & 88.00 & 73.33 & 84.62 & 90.32 & 84.85 \\ \hline
\textbf{Threshold 0.7}  \\
ViT-B-32      & 80.77 & 68.97 & 93.33 & 44.83 & 83.87 & 68.42 \\
ViT-ShipSemVL & 88.46 & 79.31 & 80.00 & 75.00 & 83.87 & 74.29 \\ \hline
\textbf{Threshold 0.5}  \\
ViT-B-32      & 88.46 & 58.33 & 93.33 & 37.14 & 67.74 & 52.50 \\
ViT-ShipSemVL & 88.46 & 69.70 & 86.67 & 52.17 & 70.97 & 57.89 \\ \hline
\end{tabular}%
}
\end{table}

\section{Discussion}
\label{sec:dis}
We constructed a Vision-Language dataset, ShipSem-VL, and proposed an innovative framework for Semantic-Aware Ship Detection (SASD) by integrating Vision-Language Models (VLMs) with an adaptive multi-scale sliding window strategy. The construction of the ShipSem-VL dataset, with its fine-grained semantic annotations, marks a significant advancement in SASD by enabling precise detection of ship attributes in complex scenarios. Additionally, our dataset also provides a foundation for related fields, such as remote sensing image generation for specific categories.

The experimental results demonstrate the effectiveness of the proposed approach across multiple tasks in SASD. Notably, the framework exhibits significant improvements in precision, particularly for tasks involving detailed semantic descriptions. However, challenges remain, such as reduced recall in visually dense scenes due to overlapping ROIs. These findings highlight both the potential and the limitations of text-driven detection frameworks in complex remote sensing contexts. In the future, we aim to further explore the potential of semantic target detection for specific objectives using specialized datasets. The scalability of this approach can be expanded through self-supervised or weakly supervised learning techniques to reduce reliance on manual annotations. Moreover, we plan to optimize our method and framework to extend its application to a broader range of categories.
\section*{Acknowledgment}
This work was supported by the National Natural Science Foundation of China (42125503, 42430602).

\bibliographystyle{IEEEtranN}
\bibliography{references}

\begin{thebibliography}{24}
\providecommand{\natexlab}[1]{#1}
\providecommand{\url}[1]{#1}
\csname url@samestyle\endcsname
\providecommand{\newblock}{\relax}
\providecommand{\bibinfo}[2]{#2}
\providecommand{\BIBentrySTDinterwordspacing}{\spaceskip=0pt\relax}
\providecommand{\BIBentryALTinterwordstretchfactor}{4}
\providecommand{\BIBentryALTinterwordspacing}{\spaceskip=\fontdimen2\font plus
\BIBentryALTinterwordstretchfactor\fontdimen3\font minus \fontdimen4\font\relax}
\providecommand{\BIBforeignlanguage}[2]{{%
\expandafter\ifx\csname l@#1\endcsname\relax
\typeout{** WARNING: IEEEtranN.bst: No hyphenation pattern has been}%
\typeout{** loaded for the language `#1'. Using the pattern for}%
\typeout{** the default language instead.}%
\else
\language=\csname l@#1\endcsname
\fi
#2}}
\providecommand{\BIBdecl}{\relax}
\BIBdecl

\bibitem[Paolo et~al.(2024)Paolo, Kroodsma, Raynor, Hochberg, Davis, Cleary, Marsaglia, Orofino, Thomas, and Halpin]{paolo2024satellite}
F.~S. Paolo, D.~Kroodsma, J.~Raynor, T.~Hochberg, P.~Davis, J.~Cleary, L.~Marsaglia, S.~Orofino, C.~Thomas, and P.~Halpin, ``Satellite mapping reveals extensive industrial activity at sea,'' \emph{Nature}, vol. 625, no. 7993, pp. 85--91, 2024.

\bibitem[Ren et~al.(2016)Ren, He, Girshick, and Sun]{ren2016faster}
S.~Ren, K.~He, R.~Girshick, and J.~Sun, ``Faster r-cnn: Towards real-time object detection with region proposal networks,'' \emph{IEEE transactions on pattern analysis and machine intelligence}, vol.~39, no.~6, pp. 1137--1149, 2016.

\bibitem[Du(2018)]{du2018understanding}
J.~Du, ``Understanding of object detection based on cnn family and yolo,'' in \emph{Journal of Physics: Conference Series}, vol. 1004.\hskip 1em plus 0.5em minus 0.4em\relax IOP Publishing, 2018, p. 012029.

\bibitem[Radford et~al.(2021)Radford, Kim, Hallacy, Ramesh, Goh, Agarwal, Sastry, Askell, Mishkin, Clark, et~al.]{radford2021learning}
A.~Radford, J.~W. Kim, C.~Hallacy, A.~Ramesh, G.~Goh, S.~Agarwal, G.~Sastry, A.~Askell, P.~Mishkin, J.~Clark \emph{et~al.}, ``Learning transferable visual models from natural language supervision,'' in \emph{International conference on machine learning}.\hskip 1em plus 0.5em minus 0.4em\relax PMLR, 2021, pp. 8748--8763.

\bibitem[Shen et~al.(2021)Shen, Li, Tan, Bansal, Rohrbach, Chang, Yao, and Keutzer]{shen2021much}
S.~Shen, L.~H. Li, H.~Tan, M.~Bansal, A.~Rohrbach, K.-W. Chang, Z.~Yao, and K.~Keutzer, ``How much can clip benefit vision-and-language tasks?'' \emph{arXiv preprint arXiv:2107.06383}, 2021.

\bibitem[Liu et~al.(2024{\natexlab{a}})Liu, Chen, Guan, Zhou, Zhu, Ye, Fu, and Zhou]{liu2024remoteclip}
F.~Liu, D.~Chen, Z.~Guan, X.~Zhou, J.~Zhu, Q.~Ye, L.~Fu, and J.~Zhou, ``Remoteclip: A vision language foundation model for remote sensing,'' \emph{IEEE Transactions on Geoscience and Remote Sensing}, 2024.

\bibitem[Li et~al.(2023)Li, Wen, Hu, and Zhou]{li2023rs}
X.~Li, C.~Wen, Y.~Hu, and N.~Zhou, ``Rs-clip: Zero shot remote sensing scene classification via contrastive vision-language supervision,'' \emph{International Journal of Applied Earth Observation and Geoinformation}, vol. 124, p. 103497, 2023.

\bibitem[Pan et~al.(2023{\natexlab{a}})Pan, Ma, and Bai]{pan2023prior}
J.~Pan, Q.~Ma, and C.~Bai, ``A prior instruction representation framework for remote sensing image-text retrieval,'' in \emph{Proceedings of the 31st ACM International Conference on Multimedia}, 2023, pp. 611--620.

\bibitem[Pan et~al.(2024)Pan, Ma, Ma, Bai, and Chen]{pan2024pir}
J.~Pan, M.~Ma, Q.~Ma, C.~Bai, and S.~Chen, ``Pir: Remote sensing image-text retrieval with prior instruction representation learning,'' 2024.

\bibitem[Zhang et~al.(2024)Zhang, Zhao, Guo, and Yin]{zhang2024rs5m}
Z.~Zhang, T.~Zhao, Y.~Guo, and J.~Yin, ``Rs5m and georsclip: A large scale vision-language dataset and a large vision-language model for remote sensing,'' \emph{IEEE Transactions on Geoscience and Remote Sensing}, 2024.

\bibitem[Pan et~al.(2023{\natexlab{b}})Pan, Ma, and Bai]{pan2023reducing}
J.~Pan, Q.~Ma, and C.~Bai, ``Reducing semantic confusion: Scene-aware aggregation network for remote sensing cross-modal retrieval,'' in \emph{Proceedings of the 2023 ACM International Conference on Multimedia Retrieval}, 2023, pp. 398--406.

\bibitem[Li et~al.(2022)Li, Zhang, Zhang, Yang, Li, Zhong, Wang, Yuan, Zhang, Hwang, et~al.]{li2022grounded}
L.~H. Li, P.~Zhang, H.~Zhang, J.~Yang, C.~Li, Y.~Zhong, L.~Wang, L.~Yuan, L.~Zhang, J.-N. Hwang \emph{et~al.}, ``Grounded language-image pre-training,'' in \emph{Proceedings of the IEEE/CVF Conference on Computer Vision and Pattern Recognition}, 2022, pp. 10\,965--10\,975.

\bibitem[Zhong et~al.(2022)Zhong, Yang, Zhang, Li, Codella, Li, Zhou, Dai, Yuan, Li, et~al.]{zhong2022regionclip}
Y.~Zhong, J.~Yang, P.~Zhang, C.~Li, N.~Codella, L.~H. Li, L.~Zhou, X.~Dai, L.~Yuan, Y.~Li \emph{et~al.}, ``Regionclip: Region-based language-image pretraining,'' in \emph{Proceedings of the IEEE/CVF conference on computer vision and pattern recognition}, 2022, pp. 16\,793--16\,803.

\bibitem[Zhan et~al.(2023)Zhan, Xiong, and Yuan]{zhan2023rsvg}
Y.~Zhan, Z.~Xiong, and Y.~Yuan, ``Rsvg: Exploring data and models for visual grounding on remote sensing data,'' \emph{IEEE Transactions on Geoscience and Remote Sensing}, vol.~61, pp. 1--13, 2023.

\bibitem[Hu et~al.(2023)Hu, Yuan, Wen, Lu, and Li]{hu2023rsgpt}
Y.~Hu, J.~Yuan, C.~Wen, X.~Lu, and X.~Li, ``Rsgpt: A remote sensing vision language model and benchmark,'' \emph{arXiv preprint arXiv:2307.15266}, 2023.

\bibitem[Pang et~al.(2024)Pang, Wu, Li, Liu, Sun, Li, Weng, Wang, Feng, Xia, et~al.]{pang2024h2rsvlm}
C.~Pang, J.~Wu, J.~Li, Y.~Liu, J.~Sun, W.~Li, X.~Weng, S.~Wang, L.~Feng, G.-S. Xia \emph{et~al.}, ``H2rsvlm: Towards helpful and honest remote sensing large vision language model,'' \emph{arXiv preprint arXiv:2403.20213}, 2024.

\bibitem[Sun et~al.(2022)Sun, Feng, Li, Ye, Kang, and Huang]{sun2022visual}
Y.~Sun, S.~Feng, X.~Li, Y.~Ye, J.~Kang, and X.~Huang, ``Visual grounding in remote sensing images,'' in \emph{Proceedings of the 30th ACM International Conference on Multimedia}, 2022, pp. 404--412.

\bibitem[Li et~al.(2024{\natexlab{a}})Li, Wen, Hu, Yuan, and Zhu]{li2024vision}
X.~Li, C.~Wen, Y.~Hu, Z.~Yuan, and X.~X. Zhu, ``Vision-language models in remote sensing: Current progress and future trends,'' \emph{IEEE Geoscience and Remote Sensing Magazine}, 2024.

\bibitem[Ma et~al.(2024)Ma, , and Bai]{ma2024direction}
Q.~Ma, , and C.~Bai, ``Direction-oriented visual--semantic embedding model for remote sensing image--text retrieval,'' \emph{IEEE Transactions on Geoscience and Remote Sensing}, vol.~62, pp. 1--14, 2024.

\bibitem[Liu et~al.(2024{\natexlab{b}})Liu, Zeng, Ren, Li, Zhang, Yang, Jiang, Li, Yang, Su, et~al.]{liu2024grounding}
S.~Liu, Z.~Zeng, T.~Ren, F.~Li, H.~Zhang, J.~Yang, Q.~Jiang, C.~Li, J.~Yang, H.~Su \emph{et~al.}, ``Grounding dino: Marrying dino with grounded pre-training for open-set object detection,'' in \emph{European Conference on Computer Vision}.\hskip 1em plus 0.5em minus 0.4em\relax Springer, 2024, pp. 38--55.

\bibitem[Pan et~al.(2025{\natexlab{a}})Pan, Liu, Fu, Ma, Li, Paudel, Van~Gool, and Huang]{pan2025locate}
J.~Pan, Y.~Liu, Y.~Fu, M.~Ma, J.~Li, D.~P. Paudel, L.~Van~Gool, and X.~Huang, ``Locate anything on earth: Advancing open-vocabulary object detection for remote sensing community,'' in \emph{Proceedings of the AAAI Conference on Artificial Intelligence}, vol.~39, no.~6, 2025, pp. 6281--6289.

\bibitem[Pan et~al.(2025{\natexlab{b}})Pan, Liu, He, Peng, Li, Sun, and Huang]{pan2025enhance}
J.~Pan, Y.~Liu, X.~He, L.~Peng, J.~Li, Y.~Sun, and X.~Huang, ``Enhance then search: An augmentation-search strategy with foundation models for cross-domain few-shot object detection,'' in \emph{Proceedings of the Computer Vision and Pattern Recognition Conference}, 2025, pp. 1548--1556.

\bibitem[Liu et~al.(2019)Liu, Wang, Shao, Wang, and Li]{liu2019improving}
X.~Liu, Z.~Wang, J.~Shao, X.~Wang, and H.~Li, ``Improving referring expression grounding with cross-modal attention-guided erasing,'' in \emph{Proceedings of the IEEE/CVF conference on computer vision and pattern recognition}, 2019, pp. 1950--1959.

\bibitem[Li et~al.(2024{\natexlab{b}})Li, Zhang, Bi, Li, Li, Yu, Sun, and Wang]{li2024injecting}
C.~Li, W.~Zhang, H.~Bi, J.~Li, S.~Li, H.~Yu, X.~Sun, and H.~Wang, ``Injecting linguistic into visual backbone: Query-aware multimodal fusion network for remote sensing visual grounding,'' \emph{IEEE Transactions on Geoscience and Remote Sensing}, 2024.

\end{thebibliography}

\end{document}